# Synthetic Data Generation of Body Motion Data by Neural Gas Network for Emotion Recognition

S. Muhammad Hossein Mousavi
Cyrus Intelligence Research Ltd
Tehran, Iran
ORCID: 0000-0001-6906-2152
s.muhammad.hossein.mousavi@gmail.com

*Abstract*

In the domain of emotion recognition using body motion, the primary challenge lies in the scarcity of diverse and generalizable datasets. Automatic emotion recognition uses machine learning and artificial intelligence techniques to recognize a person's emotional state from various data types, such as text, images, sound, and body motion. Body motion poses unique challenges as many factors, such as age, gender, ethnicity, personality, and illness, affect its appearance, leading to a lack of diverse and robust datasets specifically for emotion recognition. To address this, employing Synthetic Data Generation (SDG) methods, such as Generative Adversarial Networks (GANs) and Variational Auto Encoders (VAEs), offers potential solutions, though these methods are often complex. This research introduces a novel application of the Neural Gas Network (NGN) algorithm for synthesizing body motion data and optimizing diversity and generation speed. By learning skeletal structure topology, the NGN fits the neurons or gas particles on body joints. Generated gas particles, which form the skeletal structure later on, will be used to synthesize the new body posture. By attaching body postures over frames, the final synthetic body motion appears. We compared our generated dataset against others generated by GANs, VAEs, and another benchmark algorithm, using benchmark metrics such as Fréchet Inception Distance (FID), Diversity, and a few more. Furthermore, we continued evaluation using classification metrics such as accuracy, precision, recall, and a few others. Joint-related features or kinematic parameters were extracted, and the system assessed model performance against unseen data. Our findings demonstrate that the NGN algorithm produces more realistic and emotionally distinct body motion data and does so with more synthesizing speed than existing methods.

## 1. Introduction
- **Definition and Importance**

Automatic Emotion Recognition (ER) [1-3, 65] technology has rapidly evolved over recent decades, becoming a crucial tool in enhancing Human-Computer Interaction (HCI) [3, 64]. This field uses advancements in artificial intelligence and machine learning to analyze human expressions, aiming to accurately interpret and respond to human emotions. The ER will mostly be conducted using different modalities of facial expressions [7, 8], vocal expressions [9], text [10], physiological signals [2, 13], and body motion [11, 12]. Emotion recognition is important because of its diverse applications across various sectors. In healthcare, it assists in monitoring patient well-being and mental health [4]. In the automotive industry, emotion detection systems improve safety by assessing the driver's alertness and emotional state [5]. Additionally, in customer service, it enables more responsive and tailored interactions, enhancing user experience and satisfaction [6]. The integration of emotion recognition technologies into everyday devices and services emphasizes its growing significance in augmenting human interactions with digital systems, making them more intuitive and empathetic.

- **Challenges**

The field of emotion recognition, particularly when using body motion or tracking data, faces significant challenges due to the scarcity of available diverse datasets [14-16]. This rarity stems from various factors, including the complexity and cost associated with capturing detailed and accurate movement data. Moreover, the use of body motion data in emotion recognition is further complicated by its inherent variability [17, 2], which results in a lack of data diversity and data scarcity of this modality, especially in emotion recognition.

The lack of data diversity means that they don't cover different variants of age, gender, ethnicity, and disability, which don't make fair and unbiased datasets. This phenomenon is sometimes called a lack of data heterogeneity.

- **Body Motion**

Body motion is the tracking of the position and rotation of various body joints across 3-Dimensional (3-D) space over time, as captured in sequential frames [18, 19]. This data is crucial for understanding the dynamic and expressive nature of human gestures and movements, which can vary significantly as emotions are expressed physically. Individuals express emotions through body movements in highly subjective ways, influenced by cultural, social, and personal differences [17], which introduces a layer of complexity in designing universally applicable recognition systems.

- **Synthetic Data Generation**

Generative AI [30] is a subset of AI where the systems learn underlying patterns of data to create new content that mimics real-world data across various formats. It is the exact opposite of discriminative AI-like classification, which solves from data to label. That means generative AI solves form labels to multiple desired data. This technology powers diverse applications, including creating new images, generating text, converting text to images, synthesizing music, producing videos, crafting digital art, translating text into music, and more. Basically, it produces real-world-like synthetic data, which its application is formally known as Synthetic Data Generation (SDG) [20-23]. The SDG emerges as a potent solution to the above-mentioned challenges, offering a way to create large, diverse, and controlled datasets that can significantly enhance the training and performance of machine learning models in emotion recognition. This approach not only addresses the issue of data scarcity but also enables the exploration of nuanced emotional states that may not be well-represented in available real-world datasets. By using algorithms capable of simulating realistic human emotions and movements, researchers can generate data that closely mimic real-life scenarios, thus providing valuable resources for training more robust and accurate emotion recognition systems. The potential of synthetic data extends beyond filling gaps in existing datasets; it also supports the development of more ethical AI systems by reducing the reliance on personal data and adhering to privacy concerns, making it a cornerstone for future advancements in the field.

- **Neural Gas Network**

The Neural Gas Network (NGN) [24] is a machine learning algorithm designed to learn topologies in an unsupervised manner, similar to Self-Organizing Maps. It excels in clustering and visualizing high-dimensional data by iteratively adapting to a set of input vectors, reducing the dimensions while preserving the topological properties of the input space. This approach is particularly valuable in complex pattern recognition tasks where the underlying data structures are non-linear and multidimensional. The NGNs distinguish themselves through their flexibility in forming clusters of varying densities, making them highly effective for applications that require detailed feature extraction [25], segmentation [26], and robust pattern recognition [25] capabilities across diverse datasets. Some of these applications implemented by Python are available in this[1] GitHub repository.

- **Contribution**

In addressing the challenge of synthesizing body motion data for emotion recognition due to data scarcity, this research utilizes a supervised version of the Neural Gas Network (NGN) [27] to generate diverse and realistic datasets. The NGN effectively maps the complex topology of the human skeletal structure by positioning its neurons, or gas particles, on critical joints of the body. These particles capture the dynamic changes in position and rotation across the 3-D space over time, creating a faithful representation of human motion. More explanation of the method is in section three.

---

[1] https://github.com/SeyedMuhammadHosseinMousavi/Neural-Gas-Network-Toolbox

- **Evaluations Materials**

In order to evaluate the robustness of our approach, some benchmark State of the Art (SoA) algorithms are considered for generating synthetic body motion datasets, which are Generative Adversarial Networks (GANs) [30], Variational Auto Encoder (VAEs) [31], Long Short-Term Memory (LSTM) [32], diffusion models [34], and copula models [33]. Also, multiple joint-related features (kinematic parameters) are extracted and fused for the machine learning evaluation aspect. Furthermore, metrics of Fréchet Inception Distance (FID) [59], Diversity [59], Fidelity [60], Dynamic Time Wrapping (DTW) [61], Mean Per-Joint Position Error (MPJPE) [50], accuracy [62], precision [62], recall [62], f1 score [62], and Matthews Correlation Coefficient (MCC) [62, 63] are employed for evaluating and comparing the quality of the generated body motions.

- **Research Questions**

This research is trying to answer the following Research Questions (RQs). RQ1: How properly does the supervised Neural Gas Network (NGN) capture and reproduce the complex dynamics of human emotional expressions through body motion compared to other generative models? RQ2: What are the key factors influencing the generalizability and diversity of body motion data generated by NGN, and how do these factors affect the emotional clarity of the synthesized outputs? RQ3: How do variations in the training dataset size and composition impact the NGN's ability to generalize body motion for unseen emotional expressions in terms of kinematic parameters? RQ4: In practical applications, how can NGN-synthesized body motion data enhance real-time emotion recognition systems? RQ5: Can the Neural Gas Network effectively differentiate and synthesize body motion data for subtle emotional variations, and what impact does this have on the model's training efficiency and data generation speed?

- **Paper Structure**

The paper is structured as follows: The introduction outlines the importance of emotion recognition using body motion data, emphasizing challenges like data scarcity and the potential of synthetic data generation with the Neural Gas Network (NGN). The "Chronicle of Prior Research" section reviews existing literature, highlighting gaps in current methodologies. The "Proposed Method" explains the application of NGN in synthesizing realistic body motion datasets. "Evaluation and Results" assesses the effectiveness of this approach through various metrics, ending with a discussion for answering RQs. Finally, the "Conclusion and Future Works" discusses the findings and proposes directions for further research.

## 2. Chronicle of Prior Research

This section mentions widely used core algorithms which has applications in SDG and covers SDG research on non-body motion modalities and body motion modality.

- **Benchmark Core Algorithms**

First, we mention the main algorithms in the field of SDG. Widely used benchmark algorithms that could be used in SDG applications are divided into unintelligent and intelligent algorithms. Unintelligent (not learning-based) methods are basic augmentation (rotation, resizing, smoothing, and more), Principal Component Analysis (PCA) [28], and Synthetic Minority Over-sampling Technique (SMOTE) [29]. PCA is a statistical technique that reduces the dimensionality of data by finding the most significant features that capture the maximum variance or information. SMOTE is a method used to generate synthetic samples from the minority class to balance the class distribution in datasets, improving the performance of classification algorithms on imbalanced data. Additionally, the most widely used intelligent (learning-based) algorithms are GANs [30], VAEs [31], LSTM [32], copula models [33], diffusion models [34], transformers [35], Convolutional Neural Network (CNN) [36]. A GAN is a type of machine learning model that involves two neural networks, a generator and a discriminator, which compete against each other to generate new, synthetic instances of data that are indistinguishable from real data. A VAE is a type of neural network that uses probabilistic encoders and decoders to model the data distribution and generate new data points by learning a latent space representation of the input data. An LSTM is a type of recurrent neural network designed to remember information for long periods of time, making it effective for tasks involving sequential data like time series prediction and natural language processing. Copula models are statistical tools used to describe the dependency structures between multiple variables, allowing modeling of their joint distribution by specifying the marginal distributions separately from the copula, which captures the dependencies. Transformers are a type of neural network architecture that relies on self-attention mechanisms to weigh the influence of different parts of the input data, enabling parallel processing and significant improvements in tasks like language understanding and translation. A CNN is a type of deep neural network that is particularly effective for processing data with a grid-like topology, such as images, using convolutional layers that apply filters to capture spatial hierarchies and features.

- **SDG Literature on Other Modalities**

This paragraph explains non-body motion SDG research. The following research [37] explores the application of Borderline-SMOTE for augmenting Electro Encephalo Graphy (EEG) data to improve emotion recognition using convolutional neural networks. The research focuses on enhancing data quality and model performance in the domain of brain-computer interfaces for emotion recognition. Also, [38] uses PCA for synthesizing tabular data similar to the probabilistic physiological data type. The following research [39] used NGN for SDG of emotional and physiological data (EEG, Electro Cardio Graphy (ECG), and Galvanic Skin Response (GSR)), increasing synthesis quality and optimizing runtime speed. Implementation can be found here[2]. The research [40] focuses on generating texts with specific sentiment/emotion labels. This method addresses challenges like the poor quality and lack of diversity in generated texts by implementing a mixture of adversarial networks, potentially improving user interaction systems in educational and therapeutic settings. For audio modality, the [41] combines a VAE with a GAN to perform emotional speech conversion. The VAE component is used to encode emotional nuances from speech, which are then enhanced for clarity and realism by the GAN component, synthesizing new vocal expressions. The following research [42] employs an LSTM to extract emotions from text, which then guides a GAN to modify facial expressions in images accordingly. The LSTM processes the textual input to determine the desired emotional expression, and the GAN alters the facial image to match this emotion. The [43] presents a method for generating synthetic cardiac signals by employing a copula-based statistical approach. The authors utilize marginal distributions of key parameters and their linear correlations to define a Gaussian copula. From this copula, random sets of parameters are generated, which are then used to synthesize cardiac signals. This study [44] addresses the task of altering facial expressions in images using conditional diffusion models. The authors propose incorporating a semantic encoder to guide the diffusion process, enabling the generation of specific emotional expressions while preserving the individual's identity. The research introduced in [45] integrates transformer models into a cycle GAN framework to enhance emotional voice conversion. The study investigates the transformer's ability to capture intra-relations among frames by augmenting the receptive field of models, aiming to improve the conversion of emotional expressions in speech. Also, for SDG using CNN, authors of [46] used the CNN algorithm for synthesizing emotional ECG signals on the DREAMER database[3], generating a realistic and diverse dataset, and answering data scarcity challenges.

- **SDG Literature on Body Motion Modality**

This paragraph explains body motion SDG research. The [48] introduces a generative adversarial network GAN-based technique for augmenting motion capture datasets with synthetic data. They introduce a novel approach to augmenting motion capture datasets using GANs. This technique enhances the diversity and realism of biomechanical data, which is crucial for improving motion analysis and other related applications. Another example that uses GAN for motion synthesis is [49], which introduces a novel framework for synthesizing diverse and high-quality human motions from a limited set of examples. Also, examples of motion synthesis using Conditional Variational Auto-Encoder (CVAE) are introduced in [50, 51]. The framework in [50] addresses various tasks such as motion prediction, completion, interpolation, and spatial-temporal recovery by treating inputs as masked motion series. It estimates parametric distributions from these inputs to generate full motion sequences and incorporates Action-Adaptive Modulation (AAM) to adjust motion styles based on action labels, enhancing the realism and coherence of the synthesized motions. Additionally, [52, 53] are Principle Component Analysis (PCA)-based motion synthesis researches which are in the category of traditional yet effective techniques. Lots of research has been conducted on different variations of LSTM for motion synthesis as they can handle sequential data (motion frames) by capturing the temporal dynamics in human motion. Their architecture allows them to remember information for long periods, making them ideal for predicting future frames in a motion sequence based on past data. This capability enables the generation of smooth, continuous motions that can adapt over time. Some of these researches are [54, 55]. The following research [57] uses copula models for hand motion synthesis. This would be used to enhance the accuracy and robustness of the motion recognition system by effectively capturing the complex, multidimensional relationships in hand movement data. The [56] presents a novel framework utilizing denoising-diffusion processes for synthesizing human motion, effectively balancing the trade-offs between motion diversity and fidelity. The framework employs a Transformer-based architecture, enabling the synthesis of long, coherent, and contextually accurate human motions while accommodating diverse control signals such as text and music. By pretraining the model as a diffusion model, MoFusion enhances its ability to complete and integrate motions, thereby improving the generation of detailed and dynamic human activities across various scenarios. The paper [58] develops a Transformer-based Variational Autoencoder (VAE) to generate realistic and variable-length 3D human motions conditioned on specific actions without needing a starting pose. This approach, termed ACTOR, utilizes action-aware latent representations to enhance action recognition and perform motion denoising, demonstrating significant advancements over prior methods on multiple datasets. Notable research which employed a three-layer CNN and a one-dimensional convolution autoencoder for motion synthesis is conducted in [47]. They utilize CNN to synthesize realistic 3D human motion. This methodology processes motion capture data to generate motion sequences that are visually smooth and authentic, employing a one-dimensional convolution autoencoder to enhance the synthesis and apply various physical constraints. All mentioned researches have advantages and disadvantages, but none guarantees the diverse synthesized body motion alongside high runtime speed (low complexity), which our method addresses in the next section.

---

[2] https://github.com/SeyedMuhammadHosseinMousavi/Synthetic-Data-Generation-by-Supervised-Neural-Gas-Network
[3] https://zenodo.org/records/546113

### 3. Proposed Method
- **NGN on Body Motion**

Employing the NGN algorithm for synthesizing body motion in emotion recognition is conducted for the first time in this research. As mentioned in the introduction, the NGN's main purpose is to learn the topology of any input data. Here, the input data is the human body skeleton structure in which NGN learns its topology by scattering gas particles (neurons) over body joints. The process begins by learning from existing datasets to understand the typical movement patterns associated with different emotions. The NGN adapts to these patterns, with each neuron adjusting its position in the 3-D space to minimize the distance to the nearest data points (joints), which represent specific postures. Over iterations, this creates a finely tuned network that can interpolate new body motions by generating intermediate postures, effectively filling gaps in the data landscape. Once the network has learned the typical postural dynamics, it generates new sequences of body motion by simulating the transition of these gas particles across frames. This results in the synthesis of new body motions that maintain the inherent emotional expressions of the original data while introducing novel variations. These variations come from positioning gas particles around random positions of body joints due to algorithm parameters and learning processes. So, determining the number of gas particles (neurons) and parameter variables plays a significant role in the final synthesized body motion. An example of bad parameter tuning would be considering high numbers of iterations and matching the number of joins and gas particles, which ends up being almost identical between input and output body motions. Also, considering the low number of iterations and gas particles (less than body joints), it ends up with deformed body motion.

- **Employed Data Type**

We used the BioVision Hierarchy (BVH) data type[4] in our experiments. The BVH files are a standard format used to store motion capture data for humans and other figures. These files encapsulate movement data by defining a hierarchical skeleton structure composed of joints, along with the animation data specifying the motion of each joint. Each joint is defined by its name, position in 3-D space, and rotation (typically in degrees), which dictates how the joint moves relative to its parent joint in the hierarchy. The structure of a BVH file is split into two main sections: the hierarchy section and the motion section. The hierarchy section defines the skeletal structure, detailing the connections between joints (such as 'Hip', 'Knee', 'Ankle') and their initial positions and channels of rotation (e.g., 'Xrotation', 'Yrotation', 'Zrotation'). The motion section contains frame-by-frame animation data, specifying the rotation of each joint and the position of the root joint at each point in time[5]. Figure 1 depicts a BVH sample from the human body.

---
[4] https://research.cs.wisc.edu/graphics/Courses/cs-838-1999/Jeff/BVH.html

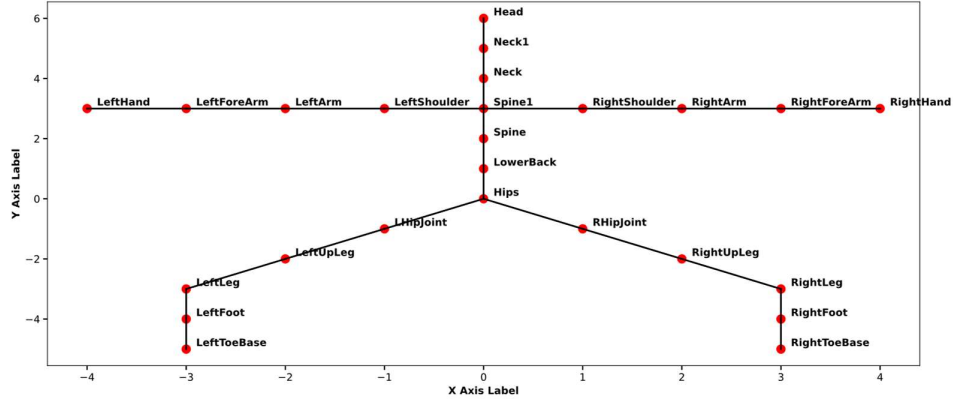

Figure 1. Visual representation of a BVH sample structure of the human body

- **Methodology Theoretical Implementation**

NGN generally consists of two layers (input and competition). The input layer receives the input features and connects each feature to every neuron in the next layer. The competition layer is where neurons compete to be closest to the input vector. The second layer comprises neurons that compete to be closer to the input data through a process that adjusts their weights. It has a couple of steps in which it computes the distances between the input vector and all neurons and ranks neurons based on distance. Then, it updates neuron weights based on their rank (by a neighborhood function). Finally, it adjusts the learning rate and neighborhood size over time to stabilize the network, ending in convergence. There are no hidden layers as found in more traditional neural networks. The NGN equational structure is as follows. Gas particles or neurons:

$$\boldsymbol{w}_i = \text{Random(input\_dim) for } i = 1 \text{ to } N \tag{1}$$

Neurons are initialized randomly in the input space, where they will later adapt to the structure of the data. Distance calculation:

$$d(\boldsymbol{v}, \boldsymbol{w}_i) = \|\boldsymbol{v} - \boldsymbol{w}_i\| \tag{2}$$

The Euclidean distance between the data vector $\boldsymbol{v}$ and each neuron $\boldsymbol{w}_i$ is calculated. This determines how close each neuron is to the current data vector. Ranking:

$$\text{rank}(\boldsymbol{v}, i) = \sum_{k=1}^{N} \mathbf{1}\big(d(\boldsymbol{v}, \boldsymbol{w}_k) < d(\boldsymbol{v}, \boldsymbol{w}_i)\big) \tag{3}$$

Neurons are ranked based on their distance to the data vector. This ranking influences the magnitude of the update each neuron receives. Neurons with a higher rank (closer to the data vector) receive larger scores and updates. Neuron update:

$$\boldsymbol{w}_i \leftarrow \boldsymbol{w}_i + \epsilon_t \cdot \exp\left(-\frac{\text{rank}(\boldsymbol{v},i)}{\lambda_t}\right) \cdot (\boldsymbol{v} - \boldsymbol{w}_i) \tag{4}$$

Neurons are updated based on their rank and proximity to the data vector. Neurons closer to the input data receive larger updates, allowing them to converge toward regions of high data density. Learning Rate (Epsilon) $\epsilon_t$ is a parameter that controls the magnitude of the updates made to each neuron. It decreases over time, allowing the network to stabilize as it converges. Neighborhood Range (Lambda) $\lambda_t$ determines the extent of influence that a data vector has on nearby neurons. It verifies that not only the closest neuron is updated but also those in its neighborhood.

---
[5] https://mathematica.stackexchange.com/questions/60292/how-to-build-a-bvh-a-motion-capture-file-format-player-in-mathematica

After explaining the NGN core implementation, the following would explain the implementation of NGN in forming the body motion BVH sample. Neurons are $w_i$ which are joint candidates, including both positions and rotations. Data vector $x$ is a single frame from the BVH file, representing the positions and rotations of all joints. Learning rate (Epsilon) $\epsilon_t$ controls how much the neurons adjust toward the current joints in the frame during training. Neighborhood range (Lambda) $\lambda_t$ certifies that not only the closest neuron to the joint but also those nearby are updated. (The extent of influence). $Rank\ (x, i)$ determines how closely a neuron matches the input. Neurons with a higher rank (closer to the joint) receive larger updates. Neurons are initialized with random positions and rotations for each joint in the body. This sets the starting point for the adaptation process.

$$w_i = [pos_{1i}, rot_{1i}, \dots, pos_{Ji}, rot_{Ji}] = \text{Random(input\_dim)} \quad \text{for } i = 1 \text{ to } N \quad (5)$$

The distance between joints of the current frame $x$ and each neuron $w_i$ is calculated, incorporating both positional and rotational differences.

$$d(x, w_i) = \sqrt{\sum_{j=1}^{J} \left( \|pos_j - pos_{ji}\|^2 + \|rot_j - rot_{ji}\|^2 \right)} \quad (6)$$

Neurons are adjusted based on how close they are to the input frame. This process ensures that neurons gradually converge toward configurations that represent realistic body poses, including both positions and rotations.

$$w_i \leftarrow w_i + \epsilon_t \cdot \exp\left(-\frac{\text{rank}(x,i)}{\lambda_t}\right) \cdot (x - w_i) \quad (7)$$

After training, the closest neurons to joints are used to generate new frames, with small variations introduced by adding Gaussian noise. This creates new motion data that is similar but not identical to the original.

$$\text{new\_frame} = w_{\text{closest}} + \mathcal{N}(0, \sigma^2) \quad (8)$$

Finally, combined generated frames create an output motion capture for the input body motion.

$$\text{Full Motion Sequence} = \sum_{t=1}^{T} \text{new\_frame}_t \quad (9)$$

Figure 2 depicts the visual performance of the NGN algorithm for filling the topology of the skeleton structure in two rows in different selected frames and with different numbers of gas particles. Also, Figure 3 illustrates generated body motion results from GAN and NGN compared with the original on a specific frame.

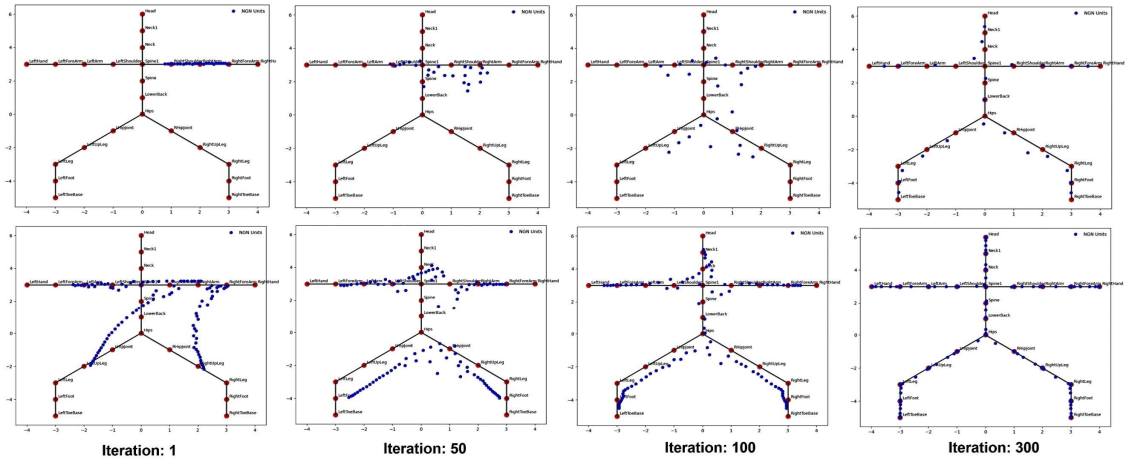

Figure 2. NGN performance with different gas particles on different iterations. Top, 24 gas particles, and bottom 100.

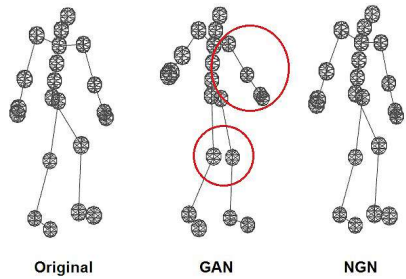

Figure 3. Comparing GAN and NGN-generated body motion with the original sample (depressed sample)

Looking at Figure 2, in the first row, the slow convergence of gas particles is vivid compared with the second row, which has 100 gas particles. It is visible that with 100 gas particles, the partial convergence starts around iteration 50, but it is not the same case with 24 gas particles. Also, at the end of the iteration, more gas particles are available to form the final body motion with 100 gas particles, which could increase diversity and open our hand for joint selection. By looking closely at Figure 3 and the GAN-generated sample, knee joints are closer to each other, which is more correlated with neutral emotion rather than depression, as the original emotion is depression. Furthermore, open hands are correlated with proud and happy emotions, not depression. The NGN-generated sample is more similar to the original sample but with some variation, which is what we are looking for.

## 4. Evaluation and Results

- **SoA Algorithms on Body Motion Synthesis**

This section starts with discussing how unintelligent and intelligent SDG methods generate BVH body motion files. Non-learning base and learning base algorithms generate body motion modality differently. For instance, in basic augmentation, rotation will be applied to body joints to create new motion sequences by changing the angle of joints, resizing will adjust the scale of motion to simulate different body sizes, and smoothing can help in reducing noise in motion data, making it more fluid and realistic. PCA can be used in SDG by identifying the principal components of body motion data in BVH files. This reduction helps in isolating the most significant motion patterns, which can then be exaggerated or combined to generate new, plausible motion sequences. By manipulating these principal components, new variations of the original motion data that maintain the core characteristics but differ in specific aspects can be synthesized. SMOTE will be utilized to address class imbalance in motion recognition datasets by generating synthetic examples of underrepresented body motions. By interpolating between similar instances in the minority class, SMOTE will create new, synthetic BVH files that help balance the dataset, improving the robustness and performance of emotion recognition models trained on this data.

GANs will generate new and diverse body motions by training two models: a generator that creates motions and a discriminator that evaluates them. The generator will learn to produce BVH files that mimic authentic human motions, convincing the discriminator that these generated motions are real, thus enriching the dataset with varied and realistic body movements. VAEs will model the distribution of existing body motions in the dataset and generate new data points by sampling from this learned distribution. This process involves encoding motion data into a latent space and then decoding from this space to create new BVH files, effectively generating new motion sequences that are variations of the learned data. LSTMs will capture the temporal dynamics and dependencies in body motion data, allowing the generation of sequential motion data that follows logical and temporally coherent progressions. This makes it particularly effective for creating longer sequences of body motion that are realistic and maintain consistency in movement patterns. Copula models will effectively capture and recreate the complex dependencies between different joints and motions in body motion data. By modeling the joint distribution of these variables separately from their margins using copulas, new synthetic motion data that respects the inherent dependencies observed in real data will be generated, ensuring that the synthetic motions are both plausible and varied. Diffusion models will transform a simple noise distribution into a complex distribution of body motions through a gradual denoising process. This approach will allow for the generation of high-quality, realistic body motions by iteratively refining random noise based on learned data characteristics, producing new BVH files that exhibit natural human movement patterns. Transformers will use their self-attention mechanisms to analyze and generate body motion data by considering the entire sequence of movements at once. This allows them to understand and replicate complex interactions between different body parts over time, generating new sequences that maintain logical and contextually appropriate motions and enhancing the realism and diversity of synthetic BVH data. CNNs will process body motion data by applying convolutional filters that detect spatial hierarchies and dependencies in motion patterns. This capability will enable them to generate new motion data by capturing and emphasizing characteristic features in body movements, creating synthetic BVH files that are realistic and applicable for training robust emotion recognition models.

- **NGN Algorithm on Body Motion Synthesis**

The NGN is trained on normalized motion data for each class to adapt the positions of its neurons to represent the data distribution effectively. When generating new samples, the process involves randomly selecting a neuron from this trained network and adding Gaussian noise to create variability. This method ensures that each generated sample is a novel instance, influenced by the overall characteristics of the class data rather than being directly derived from any specific existing sample. The generation of three new samples per class is achieved by repeating this process independently, using the diversity within the network's neurons to produce varied and representative synthetic motion data.

In the configuration for the Neural Gas Network (NGN) used for generating synthetic motion data, several parameters guide the training and generation process. The network is set up with 50 neurons, which allows for capturing a broad range of patterns within the motion data across 50 iterations, facilitating detailed learning over a moderate number of steps. The learning rate parameters, epsilon_initial and epsilon_final, start at 0.3 and decay to 0.05, controlling the adaptation speed of the neurons to the data, with a gradual slowdown in adjustments as learning progresses. Similarly, lambda_initial and lambda_final set the influence range from 10 to 0.1, decreasing the neighborhood effect over time and allowing finer local adjustments in the later stages of training. The system is configured to generate five new samples per class, introducing variations through Gaussian noise with a standard deviation of 3.0, which smoothes the generated data to simulate realistic motion sequences by softening abrupt changes and enhancing the natural flow of movements.

- **The Dataset**

We used a body motion dataset introduced in [66] called "100Style". The "100Style" dataset contains over 4 million motion capture frames, detailing 100 diverse locomotion styles for real-time animation style modeling. Captured at 60 frames per second within a 4.5m x 4.5m space, the dataset employs Xsens technology to record full-body motion data across 28 joints from a single 182 cm tall actor, excluding finger transformations. The comprehensive dataset includes various movements such as idling, walking, and running, each rigorously categorized into multiple gait types like backward and forwards, runs and walks, sidestepping, and transitions. Additionally, detailed data processing techniques such as mirroring for symmetric styles and extensive feature extraction (trajectory positions, joint positions, velocities, and rotations) enhance the dataset's utility for developing systems that dynamically modulate motion styles based on user input. The dataset is available at [this link][6]. As this paper is about synthesizing emotional body motion data, just four style folders of angry, depressed, neutral, and proud, which included emotional body movement, are selected. In total, 32 samples are selected, including all actions. The selected body motion files are in different sizes, but the average number of frames is 3000. The selected data is considered balanced, not biased, regarding sample distribution across classes, as each class contains eight samples. This balance helps in training models more effectively, as it reduces the likelihood of the model overfitting to overrepresented classes or underfitting to underrepresented ones. Balanced datasets provide an equal opportunity for learning each class's characteristics, leading to potentially better generalization and performance on unseen data. Figure 4 illustrates the trajectories of body motions for different emotional states of 32 selected original and 40 synthetic samples generated by the proposed method captured in a two-dimensional plot of X and Z positions. This visualization provides a compelling insight into how emotional states can influence the walking movement behavior of individuals.

---

[6] https://www.ianxmason.com/100style/

The figure shows that synthetic samples resembled the original baseline samples but with diverse variations, keeping the baseline walking pattern.

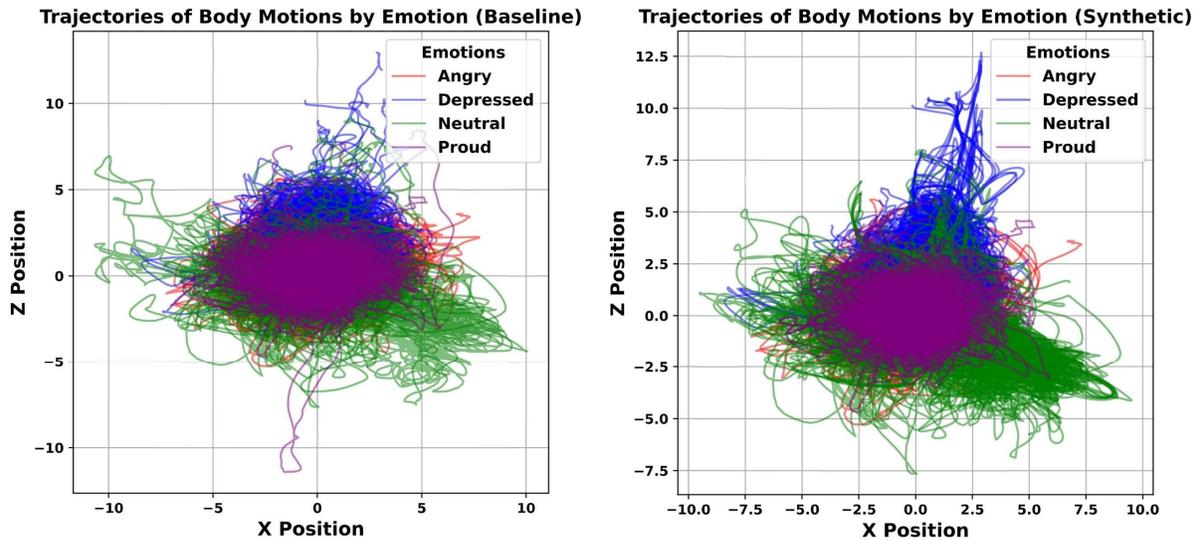

Figure 4. Walking trajectories of 32 original samples (left) and 40 NGN-generated synthetic samples of all emotions

- **Data Processing and Features**

In order to evaluate the performance of our proposed method and comparison, we manually process the data. The system reads the raw data for each class and then parses them to their structural elements. The parsing involves reading and interpreting the data structure that describes the hierarchical skeleton of the body, including joint names, their relationships, and the motion data frames, which detail the rotation and possibly translation of each joint over time. The process typically splits into handling the hierarchical skeleton definitions in the HIERARCHY section and sequentially applying the motion data from the MOTION section to animate the skeleton. After parsing, we interpolated data frames to have a unique number of frames for input and output body motions. Also, a smoothing Gaussian function is considered to lower the joint jittering ratio as postprocessing for the body motion synthesizing section. Standardization is the next step, as standardizing data is crucial in machine learning to confirm each feature contributes equally, preventing bias towards features with larger scales. Processed samples are now ready for feature extraction.

Multiple joint-related features (kinematic parameters) are extracted and fused for the evaluation. Kinematic parameters [69] or joint-related features are various metrics, features, and quantities used to describe body joints in the 3-D space. These parameters are crucial in understanding how a body or its parts move through space over frames. Kinematic parameters are specifically designed to analyze body motions across frames or sequences, typically resulting in greater accuracy compared to statistical features. Additionally, having a higher count of relevant features positively influences training, giving kinematic parameters an edge in this context. We extracted ten kinematic parameters, namely, velocity, acceleration, jerk, angular velocity, range of motion, spatial path, and harmonics magnitude.

Velocity measures the speed at which a joint moves in 3-D space. In emotion recognition, higher velocities are associated with more dynamic or intense emotions like anger, whereas slower velocities could indicate depression or a neutral state. Acceleration quantifies the rate of change of velocity. Sudden changes in acceleration have correlated with sudden emotional bursts, such as the rapid onset of anger or a proud exclamation. Conversely, minimal acceleration is more common in depressed or neutral states. Jerk measures the rate of change of acceleration, providing insight into the fluidity or abruptness of movement. High jerk values are seen in angry motions, which are typically quick and abrupt, while lower jerk could relate to smoother movements seen in states of pride or neutrality. Angular velocity describes the speed of rotation around joints. Rapid angular velocity is linked to emotions that involve pronounced body rotations, such as pride or anger, while slower angular velocities are indicative of depression or neutral states. Range of motion refers to the extent of movement at a joint. A wide range of motion is associated with pride, which often involves expansive gestures, or anger, which includes large, emphatic movements. A limited range of motion is visible in depressed individuals. The spatial path is the trajectory a joint follows through space and can also provide emotional cues. Complex or erratic paths are associated with anger, while straightforward, minimal deviation paths are more typical of depression or neutrality. Harmonics magnitude measures the intensity of the oscillatory components of movement. High harmonics are present in the expressive movements associated with pride or anger, whereas low harmonics suggest subdued emotions like depression.

The kernel density plots in Figure 5 illustrate the distribution of various kinematic features—velocity means, acceleration means, jerk mean, angular velocity means, range of motion, and spatial path mean—across different emotional states: angry, depressed, neutral, and proud for the mix of baseline and synthetic samples. These distributions show that the proud emotion typically exhibits slightly higher velocities, suggesting more vigorous movement, whereas the angry emotion is characterized by higher acceleration and jerk, indicating more rapid and abrupt movements. In contrast, depressed and neutral emotions tend to show lower values in these parameters, reflecting slower and smoother movements. This suggests that emotional states significantly influence the dynamics of body movements, with each emotion manifesting distinct physical characteristics in terms of speed and motion fluidity. The correlation heatmap depicted in Figure 6 provides insight into the relationships between various kinematic features (baseline and synthetic). Notably, acceleration_mean and spatial_path exhibit a strong positive correlation (0.90), indicating that increases in acceleration are closely associated with longer spatial paths. Similarly, velocity_mean and acceleration_mean are also highly correlated (0.90), suggesting that higher velocities tend to occur alongside higher accelerations. Conversely, angular_velocity_mean shows a significant negative correlation with range_of_motion (-0.68), implying that higher angular velocities may correspond to more restricted movements. Figure 6 effectively highlights the interdependencies between motion-related features, enhancing our understanding of how different aspects of movement are related in body motion analysis.

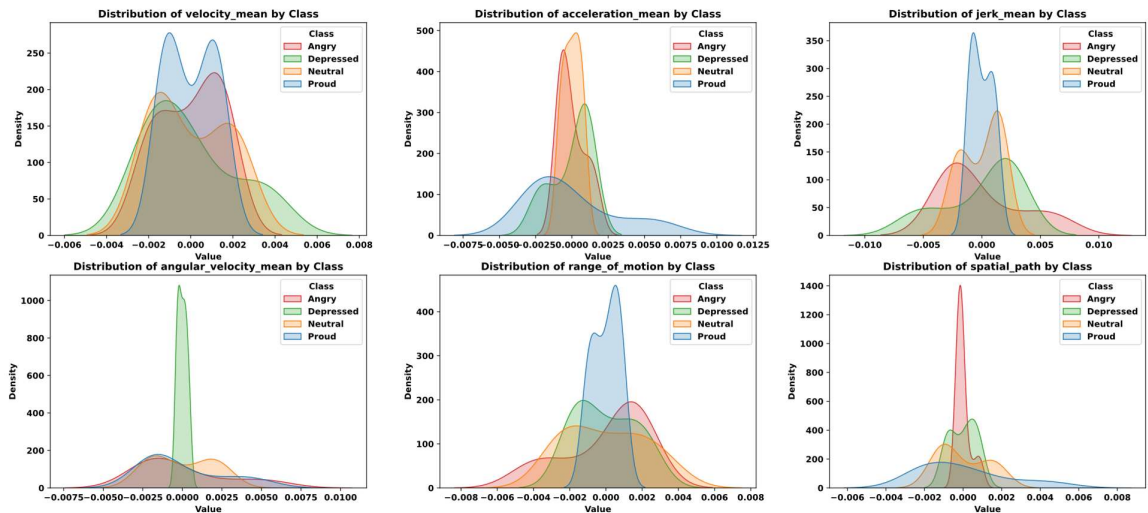

Figure 5. Kernel density plots showing the distribution of kinematic features (velocity, acceleration, jerk, angular velocity, range of motion, and spatial path) across four emotional states: angry, depressed, neutral, and proud (baseline and synthetic)

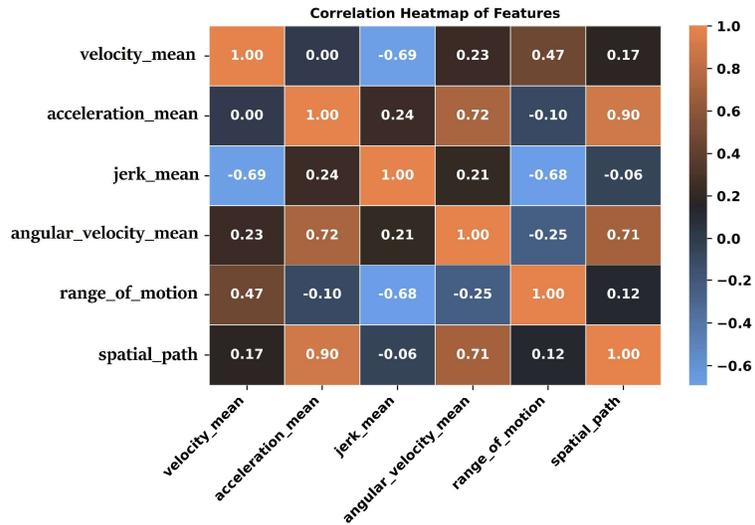

Figure 6. Correlation heatmap of kinematic features illustrating positive and negative relationships (baseline and synthetic)

- **Metrics and Classifier**

Some of the most well-known metrics in the body motion era are considered for the evaluation process as follows. Fréchet Inception Distance (FID) [59] measures the quality of synthetic data by comparing the distribution of generated body motion sequences to real body motion sequences. This is done by extracting feature vectors from both real and generated sequences using a deep learning model and then calculating the distance between these distributions. A lower FID score indicates closer similarity between the synthetic and real data distributions, suggesting a more accurate representation by the generative model. Diversity [59] is a metric that evaluates the variance within the generated data. Essentially, it measures how different the generated motion sequences are from each other, ensuring that the model produces a wide range of motions rather than repeating similar patterns. High diversity indicates that the model can generate various realistic body motions, reflecting the natural variability observed in human movements. Fidelity [60] is the accuracy with which the generated data mimics real body motions associated with specific emotions. It measures how representative the synthetic samples are of the actual data samples they are meant to emulate. High fidelity means that the generated motions are indistinguishable from real motions regarding how well they convey the intended emotions. Dynamic Time Warping (DTW) [61] is a method used to assess the similarity between two time-series sequences, such as synthetic and real body motion data. It adjusts the sequences against each other in a non-linear manner to align them optimally over time. This metric is useful for evaluating the temporal accuracy of synthetic motion sequences compared to real motion, ensuring that the timing and pacing of movements are realistically simulated. A lower DTW score indicates a better match between the synthetic and real sequences, highlighting the effectiveness of the synthetic data in replicating true human motion dynamics. Mean Per-Joint Position Error (MPJPE) [50] is a metric used to evaluate the accuracy of predicted or generated body motion data by measuring the average distance error between corresponding joint positions in the generated and the real motion data. This error is typically calculated for each joint at every frame and then averaged over all joints and frames to get the mean error. A lower MPJPE value indicates that the synthetic body motion closely matches the real human motion in terms of spatial accuracy, making it a critical measure for assessing the precision of motion capture technologies or generative models in synthesizing realistic human movements. Accuracy [62] measures the proportion of total predictions that are correct. It is a straightforward metric that tells you how often the model correctly identifies or classifies the emotion based on the generated body motion data. High accuracy indicates that the synthetic data performs well in scenarios similar to the training data but doesn't account for class imbalance. Precision [62] is the ratio of correctly predicted positive observations to the total predicted positives. It highlights the model's ability to not label a sample that is negative as positive, making it crucial for cases where the cost of a false positive is high. Recall (or sensitivity) [62] measures the ability of a model to find all the relevant cases (true positives) within a dataset. High recall means that the model captures a large proportion of positive examples, which is especially important in emotion recognition, where missing an emotional cue could be critical. The F1 score [62] is the harmonic mean of precision and recall. It is used when you need to balance precision and recall, an especially common scenario in uneven class distributions, or when false positives and false negatives have different costs. Matthews Correlation Coefficient (MCC) [62] is a correlation coefficient between the observed and predicted binary classifications. It provides a balanced measure that can be used even if the classes are of very different sizes. The MCC is generally regarded as one of the best metrics for binary classification tasks as it takes into account true and false positives and negatives.

In our research, we've found the random forest classifier [69] to be highly effective. This method is particularly suitable due to its robust handling of complex and high-dimensional data, which is typical of motion capture systems. The random forest, by building multiple decision trees and averaging their predictions, reduces the risk of overfitting and enhances the overall accuracy of the model. Due to its structure, it is easier to explain as we use feature importance to find the most important features.

- **Experiments and Results**

Out of 32 samples in the original data, eight samples were selected and completely separated for the synthesis process (two samples per class). Out of those eight samples, 10 samples for each class are generated using NGN, which creates a total of 40 synthetic samples. Testing the model is based on three combinations: baseline (22 samples), just synthetic (40 samples), and a mix of synthetic and baseline (64 samples). Also, the same number of samples are generated by other algorithms. The NGN parameters for body motion synthesis are mentioned above, but other algorithm parameters are listed as follows. GANs use a learning rate of 0.002, a beta1 value for the Adam optimizer set at 0.5, with a batch size of 16, and run for 200 epochs. VAEs use a learning rate of 0.001, a batch size of 16, with latent space dimensions set at 20 for 200 epochs. The diffusion model is configured with a learning rate of 0.001, 100 diffusion steps, a linear noise schedule, and a batch size of 16. The Gaussian copula model has 10 bins and runs for 100 epochs. LSTMs feature a learning rate of 0.001, consisting of three layers with 64 hidden units, and use a batch size of 16. Furthermore, all train and test combinations are based on 70 % training data and 30 % testing over 20 runs of Monte Carlo cross-validation.

Figure 7 depicts the NGN average error for all generated body motions of all emotional classes. These values belong to generating 10 samples from each class, bringing 40 new samples in total. The average error represents the average distance between the input samples (joint data) and their closest neurons in the network. This value indicates how well the NGN is capturing the structure of the input data, with lower values indicating a better fit.

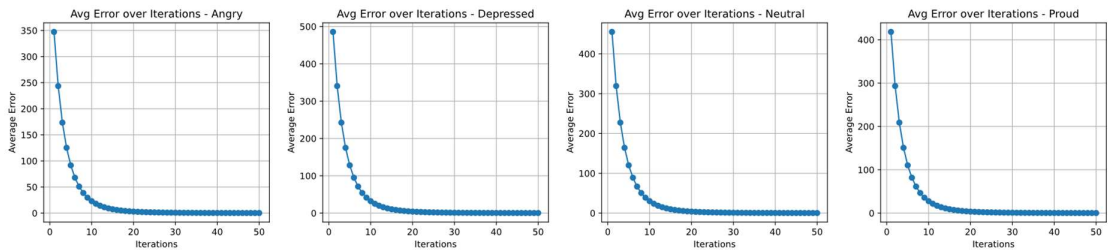

Figure 7. NGN average error for all generated body motions of all emotional classes – 40 new samples, 10 samples per class (generation runtime: 2 minutes and 42 seconds – core i7 CPU - generation 12)

Table 1 provides a detailed comparison of various generative models used for augmenting body motion data, focusing on a set of key performance metrics across different scenarios: using only synthetic data (Syn), only base data, and a combination of both (Syn+Base). Models always show improved performance metrics when synthetic and baseline data are combined. For instance, NGN achieves the highest accuracy (97%), precision (98%), and MCC (97%) in our experiment. The combination of synthetic and base data provides a richer dataset, enhancing the models' ability to learn nuanced patterns of emotion-specific movements, which in turn improves generalization and classification accuracy. The addition of synthetic data allows for a broader range of training examples, helping models to better capture the variability in human emotions expressed through body motion. This results in higher precision and recall, which is essential for reducing false positives and negatives in emotion recognition.

The diffusion model exhibits the lowest standard deviation in the Synthetic scenario (0.039), indicating highly consistent performance across different data splits. This reliability is crucial in practical applications where unpredictable performance can degrade user experience or system effectiveness. High diversity scores, such as NGN (1142 in Syn+Base) and GAN (1149 in Syn+Base), show that these models are effective at generating a wide range of motion patterns, enhancing the robustness of the emotion recognition system. Higher diversity verifies the model can handle various motion styles and intensities, which are common in real-world scenarios.

Models like NGN, GAN, and diffusion model show improved fidelity scores in Syn+Base scenarios, pointing out that the synthetic data they generate is more representative of true data distributions. This closeness to real data helps in training more accurate models, reducing the risk of overfitting to atypical, noisy, or limited training data.

The NGN model demonstrates outstanding efficiency with the shortest runtime of 2 minutes and 42 seconds and the lowest iteration, having just 22 units, which is one of our model advantages. It has to be mentioned that copula models are very optimized in that manner but not that accurate in terms of machine cleanout metrics. Also, LSTM performed the worst on most metrics. Furthermore, lower scores in FID, DTW, and MPJPE for models like the NGN, GAN, and Diffusion Model demonstrate that it is particularly effective at synthesizing data that is close to what is observed in real human movements. This capability is crucial for training systems that need to recognize subtle variations in emotional expressions conveyed through body motion.

Table 1. Evaluation results for all algorithms using all metrics – new abbreviations are: Base for Baseline, Syn for Synthetic, Gen for Generation, Conv for Convergence, Run for Runtime, Itr for Iterations, Min for Minutes, and Sec for Seconds.

| | Base | NGN | | GAN | | VAE | | Diffusion Model | | Copula Model | | LSTM | |
|---|---|---|---|---|---|---|---|---|---|---|---|---|---|
| | - | *Syn* | *Syn+Base* | *Syn* | *Syn+Base* | *Syn* | *Syn+Base* | *Syn* | *Syn+Base* | *Syn* | *Syn+Base* | *Syn* | *Syn+Base* |
| Accuracy ↑ | 80 % | **95 %** | **97 %** | 94 % | 96 % | 90 % | 92 % | 93 % | 95 % | 82 % | 85 % | 84 % | 88 % |
| Std ↑ | 0.130 | 0.044 | 0.046 | 0.096 | 0.091 | 0.064 | 0.087 | **0.039** | 0.041 | 0.254 | 0.157 | 0.111 | 0.107 |
| Precision ↑ | 86 % | **96 %** | **98 %** | 95 % | 97 % | 92 % | 94 % | 95 % | 95 % | 83 % | 86 % | 85 % | 86 % |
| Recall ↑ | 80 % | **95 %** | **97 %** | 95 % | 97 % | 91 % | 93 % | 94 % | 96 % | 82 % | 86 % | 85 % | 87 % |
| F1-Score ↑ | 79 % | **95 %** | **97 %** | 95 % | 96 % | 91 % | 93 % | 94 % | 95 % | 82 % | 85 % | 84 % | 87 % |
| MCC ↑ | 76 % | 94 % | **97 %** | **95 %** | 95 % | 91 % | 94 % | 93 % | 94 % | 84 % | 85 % | 84 % | 87 % |
| Diversity ↑ | **1162** | 1130 | 1142 | 1139 | 1149 | 1121 | 1129 | 1131 | 1138 | 1054 | 1088 | 1065 | 1095 |
| Fidelity ↑ | -0.030 | -0.014 | -0.011 | -0.013 | **-0.010** | -0.021 | -0.019 | -0.014 | -0.012 | -0.087 | -0.080 | -0.081 | -0.077 |
| FID ↓ | - | - | **3679** | - | 3680 | - | 3681 | - | **3679** | - | 3697 | - | 3691 |
| DTW ↓ | - | - | 9872 | - | **9812** | - | 9878 | - | 9873 | - | 9910 | - | 9900 |
| MPJPE ↓ | - | - | 3179 | - | **3169** | - | 3185 | - | 3180 | - | 3191 | - | 3187 |
| Gen Conv ↓ | - | **Itr 22** | - | Itr 150 | - | Itr 98 | - | Itr 110 | - | Itr 71 | - | Itr 183 | - |
| Gen Run ↓ | - | **2 Min 42 Sec** | | 25 Min | - | 15 Min | - | 11 Min | - | 3 Min 28 Sec | - | 43 Min | - |

In Figure 8, the violin plots for data reveal a range of variabilities across key performance metrics. In the first row, the moderate to wide distributions in accuracy and precision indicate inconsistencies in the baseline model's performance, particularly in handling false positives and achieving consistent, correct classifications. The narrower spread in recall shows relative stability in identifying all relevant cases, but there's room for improvement. The F1-Score and MCC, which provide balanced and holistic views of model performance, also show moderate variability, indicating the model's challenges in harmonizing precision and recall and maintaining consistency across all aspects of classification. This analysis highlights the need to refine the baseline model or its training data to enhance its reliability and effectiveness in real-world scenarios.

The violin plots for the NGN model using synthetic data demonstrate narrower distributions across most metrics compared to the baseline, indicating improved consistency and performance. Precision shows a notably wider spread, suggesting variability in the model's ability to avoid false positives consistently. However, the general tightening in distributions for accuracy, recall, F1-Score, and MCC implies that the NGN model is effectively utilizing synthetic data to enhance its overall predictive accuracy and reliability. This enhanced performance underscores the role of synthetic data in providing a more regularized and homogeneous training environment, which helps the NGN model better learn and generalize complex emotion-related patterns in body motion data.

The violin plots for the combination of synthetic and baseline data display the most favorable characteristics across all metrics, with notably tight distributions and higher medians. This indicates exceptionally consistent and high performance, with the least variability seen among the three scenarios. The combined data approach clearly enhances the model's ability to effectively balance precision and recall, as evidenced by the narrow spread in F1-Score and MCC. This superior performance is attributed to the comprehensive nature of the mixed dataset, which provides a rich variety of examples and scenarios, allowing the NGN model to achieve robust generalization capabilities. The integration of synthetic and baseline data not only stabilizes the model's output but also maximizes its accuracy and reliability in real-world emotion recognition tasks.

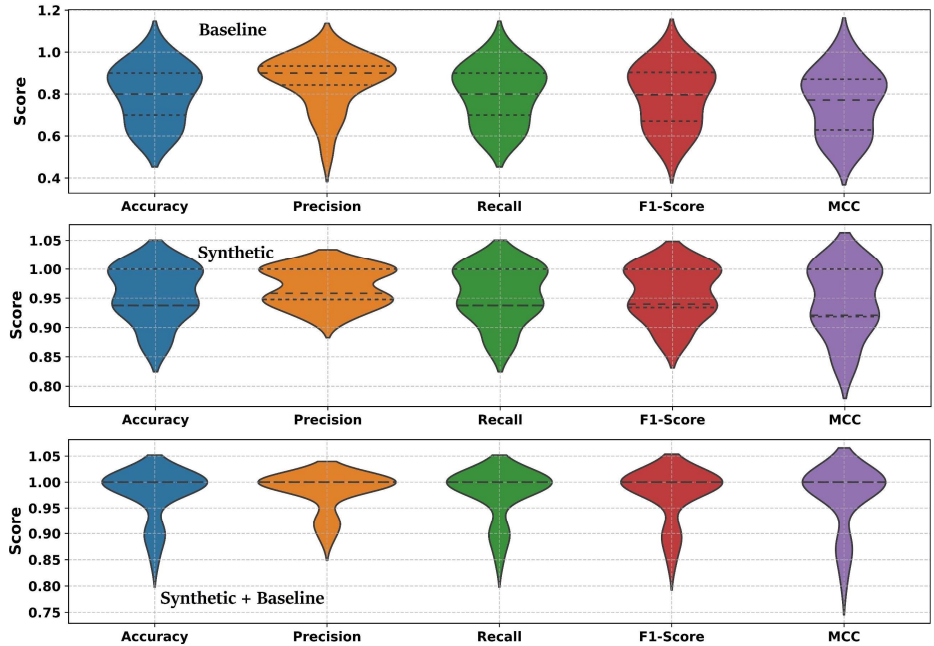

Figure 8. Violin plot for the baseline and NGN algorithm results on machine learning metrics over 20 runs – random forest classifier (train 70 %, test 30 %)

Figure 9 depicts the feature importance [68] bar plot of all emotions in the experiment (baseline and synthetic). This figure showcases the relative significance of various kinematic parameters for predicting emotional states. For the angry emotion, velocity stands out as the most significant predictor, suggesting rapid movements are characteristic of anger. In contrast, for depression, the most crucial feature is acceleration, indicating that changes in movement speed are a key indicator of depression. Neutral emotions show higher importance placed on jerk, reflecting subtle variations in speed that differentiate it from more expressive states. Finally, for proud, acceleration again appears as a significant feature, followed closely by range of motion, highlighting that both the intensity and expansiveness of movements are central to expressing pride. Implementation of the experiments can be found here[7].

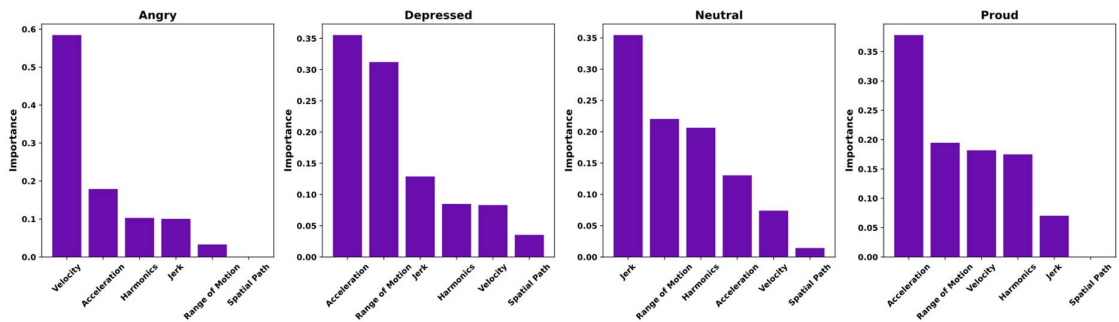

Figure 9. Feature importance bar plot for all emotions (baseline and synthetic)

---

[7] https://github.com/SeyedMuhammadHosseinMousavi/Synthetic-Data-Generation-of-Body-Motion-Data-by-Neural-Gas-Network-for-Emotion-Recognition

- **Discussion**

In the discussion, we try to answer all RQs raised in the introduction section. Regarding RQ1: The NGN-generated synthetic body motions have shown superior capability in capturing and reproducing the complex dynamics of human emotional expressions through body motion, especially when compared to baseline and other generative models like GAN, VAE, and LSTM. The metrics such as high accuracy (97%), precision (98%), and low standard deviation (0.046) in the Syn+Base setup demonstrate NGN's effectiveness in accurately modeling emotional expressions with consistency. These results are superior to those achieved with baseline or synthetic data alone, suggesting that NGN excels in integrating diverse data types to improve learning outcomes. Regarding RQ 2: Key factors influencing the generalizability and diversity of body motion data generated by NGN include the quality and variety of the training data, the model's architectural parameters, and the training regimen. The integration of synthetic and baseline data enhances both diversity (as indicated by high diversity scores) and fidelity, showcasing that these factors critically support the model in producing varied yet accurate representations of emotional expressions. This enriched data environment helps NGN data to maintain emotional clarity in the outputs, which is crucial for accurate emotion recognition. Regarding RQ 3: Variations in the training dataset size and composition significantly impact NGN's ability to generalize body motion for unseen emotional expressions. The combined data approach (Syn+Base) shows improved performance metrics across all fronts, indicating that a larger and more diverse dataset enhances the model's ability to generalize beyond the training data. This shows that both the breadth (size) and depth (composition) of the dataset are pivotal in training NGN to effectively handle new, unseen emotional expressions. Regarding RQ 4: NGN-synthesized body motion data can significantly enhance real-time emotion recognition systems by providing highly accurate and diverse training datasets that improve the model's responsiveness and adaptability. The high precision and accuracy observed in NGN outputs, especially in mixed data scenarios, suggest that NGN can offer real-time systems a robust basis for recognizing a wide range of human emotions, thereby enhancing the system's applicability and reliability in practical settings. Regarding RQ 5: The NGN is effective in differentiating and synthesizing body motion data for subtle emotional variations, as demonstrated by its high performance in precision and recall. This ability positively impacts the model's training efficiency by enabling quicker convergence on accurate models (evidenced by low iteration counts in some scenarios) and speeds up data generation (as seen with the fastest runtime among models). These capabilities make NGN particularly suitable for applications requiring a subtle understanding of human emotions through body motion, improving both the training process and the operational efficiency of deployed systems.

5. **Conclusion**

This research has demonstrated the profound capabilities of the Neural Gas Network (NGN) in synthesizing body motion data for emotion recognition, addressing significant challenges such as data scarcity and lack of heterogeneity. Through comparative analysis, NGN has proven superior in capturing and reproducing the complex dynamics of human emotional expressions through body motion when evaluated against other state-of-the-art generative models like GANs, VAEs, LSTM, diffusion models, and copula models. The integration of synthetic and baseline data has notably enhanced the NGN's performance, showcasing high precision and robust generalization across diverse emotional states. The key factors influencing the generalizability and diversity of the NGN-generated data include the quality of the synthetic data, the algorithm's inherent capacity to adapt to the complex topology of human skeletal movements, and the effective blending of synthetic and real datasets. These elements have critically improved the emotional clarity of the outputs, making the synthesized body motion data not only more diverse but also more representative of real-world variations. The use of NGN in practical applications promises significant enhancements in real-time emotion recognition systems. By generating high-fidelity, diverse body motion datasets, NGN enables these systems to operate with greater accuracy and less bias, which is crucial for applications ranging from healthcare to customer service. The ability of NGN to differentiate subtle emotional variations has also been highlighted, marking it as a potent tool for refining emotion detection algorithms and improving their responsiveness to nuanced human behaviors. Future research should explore the scalability of NGN applications in emotion recognition across larger and more varied datasets, potentially incorporating more complex emotional states and physical gestures. Further development could also focus on optimizing

NGN's computational efficiency to enhance its applicability in real-time systems where processing speed is critical. Additionally, investigating the ethical implications of synthetic data generation and its impact on privacy and data protection will be crucial as these technologies become more pervasive in everyday applications.

**REFERENCES**

[1] Picard, Rosalind W. Affective computing. MIT press, 2000.

[2] Bota, Patricia J., et al. "A review, current challenges, and future possibilities on emotion recognition using machine learning and physiological signals." IEEE access 7 (2019): 140990-141020.

[3] Mousavi, Seyed Muhammad Hossein, et al. "The Magic XRoom: A Flexible VR Platform for Controlled Emotion Elicitation and Recognition." Proceedings of the 25th International Conference on Mobile Human-Computer Interaction. 2023.

[4] Luneski, Andrej, Panagiotis D. Bamidis, and Madga Hitoglou-Antoniadou. "Affective computing and medical informatics: state of the art in emotion-aware medical applications." Studies in health technology and informatics 136 (2008): 517.

[5] Katsis, Christos D., et al. "Emotion recognition in car industry." Emotion recognition: a pattern analysis approach (2015): 515-544.

[6] Li, Xutong, and Rongheng Lin. "Speech emotion recognition for power customer service." 2021 7th International Conference on Computer and Communications (ICCC). IEEE, 2021.

[7] Mousavi, Seyed Muhammad Hossein. "Introduction to Facial Micro Expressions Analysis Using Color and Depth Images: A Matlab Coding Approach (2023)." arXiv preprint arXiv:2307.06396 (2023).

[8] Mousavi, Seyed Muhammad Hossein, and Atiye Ilanloo. "Bees Local Phase Quantisation Feature Selection for RGB-D Facial Expression Recognition." Intelligent Engineering Optimisation with the Bees Algorithm. Cham: Springer Nature Switzerland, 2024. 253-264.

[9] Zupan, Barbra, and Michelle Eskritt. "Facial and vocal emotion recognition in adolescence: a systematic review." Adolescent Research Review 9.2 (2024): 253-277.

[10] Alswaidan, Nourah, and Mohamed El Bachir Menai. "A survey of state-of-the-art approaches for emotion recognition in text." Knowledge and Information Systems 62.8 (2020): 2937-2987.

[11] Ahmed, Ferdous, ASM Hossain Bari, and Marina L. Gavrilova. "Emotion recognition from body movement." IEEE Access 8 (2019): 11761-11781.

[12] Zacharatos, Haris, Christos Gatzoulis, and Yiorgos L. Chrysanthou. "Automatic emotion recognition based on body movement analysis: a survey." IEEE computer graphics and applications 34.6 (2014): 35-45.

[13] Egger, Maria, Matthias Ley, and Sten Hanke. "Emotion recognition from physiological signal analysis: A review." Electronic Notes in Theoretical Computer Science 343 (2019): 35-55.

[14] Konak, Orhan, et al. "Overcoming Data Scarcity in Human Activity Recognition." 2023 45th Annual International Conference of the IEEE Engineering in Medicine & Biology Society (EMBC). IEEE, 2023.

[15] Lu, Yingzhou, et al. "Machine learning for synthetic data generation: a review." arXiv preprint arXiv:2302.04062 (2023).

[16] Shi, Jiaqi, et al. "Skeleton-based emotion recognition based on two-stream self-attention enhanced spatial-temporal graph convolutional network." Sensors 21.1 (2020): 205.

[17] Zhu, Wentao, et al. "Human motion generation: A survey." IEEE Transactions on Pattern Analysis and Machine Intelligence (2023).

[18] Dai, Huan, et al. "Skeletal animation based on BVH motion data." 2010 2nd International Conference on Information Engineering and Computer Science. IEEE, 2010.

[19] Meredith, Maddock, and Steve Maddock. "Motion capture file formats explained." Department of Computer Science, University of Sheffield 211 (2001): 241-244.

[20] Bauer, André, et al. "Comprehensive exploration of synthetic data generation: A survey." arXiv preprint arXiv:2401.02524 (2024).

[21] Maharana, Kiran, Surajit Mondal, and Bhushankumar Nemade. "A review: Data pre-processing and data augmentation techniques." Global Transitions Proceedings 3.1 (2022): 91-99.

[22] Mumuni, Alhassan, Fuseini Mumuni, and Nana Kobina Gerrar. "A survey of synthetic data augmentation methods in computer vision." arXiv preprint arXiv:2403.10075 (2024).

[23] Figueira, Alvaro, and Bruno Vaz. "Survey on synthetic data generation, evaluation methods and GANs." Mathematics 10.15 (2022): 2733.

[24] Martinetz, Thomas, and Klaus Schulten. "A" neural-gas" network learns topologies." (1991).

[25] Mousavi, Seyed Muhammad Hossein. "PSO Fuzzy XGBoost Classifier Boosted with Neural Gas Features on EEG Signals in Emotion Recognition." arXiv preprint arXiv:2407.09950 (2024).

[26] Mousavi, S. "Neural Gas Network Image Features and Segmentation for Brain Tumor Detection Using Magnetic Resonance Imaging Data." arXiv preprint arXiv:2301.12176 (2023).

[27] Hammer, Barbara, Marc Strickert, and Thomas Villmann. "Supervised neural gas with general similarity measure." Neural Processing Letters 21 (2005): 21-44.

[28] Pearson, Karl. "LIII. On lines and planes of closest fit to systems of points in space." The London, Edinburgh, and Dublin philosophical magazine and journal of science 2.11 (1901): 559-572.

[29] Chawla, Nitesh V., et al. "SMOTE: synthetic minority over-sampling technique." Journal of artificial intelligence research 16 (2002): 321-357.

[30] Goodfellow, Ian, et al. "Generative adversarial nets." Advances in neural information processing systems 27 (2014).

[31] Kingma, Diederik P. "Auto-encoding variational bayes." arXiv preprint arXiv:1312.6114 (2013).

[32] Hochreiter, S. "Long Short-term Memory." Neural Computation MIT-Press (1997).

[33] Sklar, M. "Fonctions de répartition à n dimensions et leurs marges." Annales de l'ISUP. Vol. 8. No. 3. 1959.

[34] Sohl-Dickstein, Jascha, et al. "Deep unsupervised learning using nonequilibrium thermodynamics." International conference on machine learning. PMLR, 2015.

[35] Vaswani, A. "Attention is all you need." Advances in Neural Information Processing Systems (2017).

[36] LeCun, Yann, et al. "Gradient-based learning applied to document recognition." Proceedings of the IEEE 86.11 (1998): 2278-2324.

[37] Chen, Yu, Rui Chang, and Jifeng Guo. "Effects of data augmentation method borderline-SMOTE on emotion recognition of EEG signals based on convolutional neural network." IEEE Access 9 (2021): 47491-47502.

[38] Park, Min-Jeong. "Synthetic data generation by probabilistic PCA." The Korean Journal of Applied Statistics 36.4 (2023): 279-294.

[39] Mousavi, S. "Synthetic Data Generation by Supervised Neural Gas Network for Physiological Emotion Recognition Data." arXiv preprint arXiv:2501.16353 (2025).

[40] Wang, Ke, and Xiaojun Wan. "Sentigan: Generating sentimental texts via mixture adversarial networks." IJCAI. 2018.

[41] Cao, Yuexin, et al. "Nonparallel Emotional Speech Conversion Using VAE-GAN." INTERSPEECH. 2020.

[42] Jubair, Mohammad Imrul, et al. "Altering Facial Expression Based on Textual Emotion." arXiv preprint arXiv:2112.01454 (2021).

[43] Pereira, Diogo Filipe, et al. "Synthesis of cardiac signals using a copula-approach." AIP Conference Proceedings. Vol. 2140. No. 1. AIP Publishing, 2019.

[44] Tang, Yihe, Wanyue Zhai, and Bihan Liu. "Facial Expression Manipulation with Conditional Diffusion Model."

[45] Fu, Changzeng, et al. "Cycletransgan-evc: A cyclegan-based emotional voice conversion model with transformer." arXiv preprint arXiv:2111.15159 (2021).

[46] Nita, Sihem, et al. "A new data augmentation convolutional neural network for human emotion recognition based on ECG signals." Biomedical Signal Processing and Control 75 (2022): 103580.

[47] Zhou, Dongsheng, et al. "3D human motion synthesis based on convolutional neural network." IEEE Access 7 (2019): 66325-66335.

[48] Bicer, Metin, et al. "Generative deep learning applied to biomechanics: A new augmentation technique for motion capture datasets." Journal of biomechanics 144 (2022): 111301.

[49] Li, Weiyu, et al. "Example-based motion synthesis via generative motion matching." ACM Transactions on Graphics (TOG) 42.4 (2023): 1-12.

[50] Cai, Yujun, et al. "A unified 3d human motion synthesis model via conditional variational auto-encoder." Proceedings of the IEEE/CVF International Conference on Computer Vision. 2021.

[51] Du, Han, et al. "Stylistic locomotion modeling and synthesis using variational generative models." Proceedings of the 12th ACM SIGGRAPH Conference on Motion, Interaction and Games. 2019.

[52] Baydal-Bertomeu, José María, et al. "A PCA-based bio-motion generator to synthesize new patterns of human running." PeerJ Computer Science 2 (2016): e102.

[53] Liu, Huajun, et al. "Human motion synthesis using window-based local principal component analysis." 2011 12th International Conference on Computer-Aided Design and Computer Graphics. IEEE, 2011.

[54] Li, Zimo, et al. "Auto-conditioned recurrent networks for extended complex human motion synthesis." arXiv preprint arXiv:1707.05363 (2017).

[55] Qiao, Yi-Ling, et al. "Learning bidirectional LSTM networks for synthesizing 3D mesh animation sequences." arXiv preprint arXiv:1810.02042 (2018).

[56] Dabral, Rishabh, et al. "Mofusion: A framework for denoising-diffusion-based motion synthesis." Proceedings of the IEEE/CVF conference on computer vision and pattern recognition. 2023.

[57] Ju, Zhaojie. A fuzzy framework for human hand motion recognition. Diss. University of Portsmouth, 2010.

[58] Petrovich, Mathis, Michael J. Black, and Gül Varol. "Action-conditioned 3d human motion synthesis with transformer vae." Proceedings of the IEEE/CVF International Conference on Computer Vision. 2021.

[59] Cervantes, Pablo, et al. "Implicit neural representations for variable length human motion generation." European Conference on Computer Vision. Cham: Springer Nature Switzerland, 2022.

[60] Wu, Yan, et al. "Saga: Stochastic whole-body grasping with contact." European Conference on Computer Vision. Cham: Springer Nature Switzerland, 2022.

[61] Boukhayma, Adnane, and Edmond Boyer. "Surface motion capture animation synthesis." IEEE transactions on visualization and computer graphics 25.6 (2018): 2270-2283.

[62] Powers, David MW. "Evaluation: from precision, recall and F-measure to ROC, informedness, markedness and correlation." arXiv preprint arXiv:2010.16061 (2020).

[63] Haji Hassani, Roushanak, et al. "Real-time motion onset recognition for robot-assisted gait rehabilitation." Journal of NeuroEngineering and Rehabilitation 19.1 (2022): 11.

[64] Sinha, Gaurav, Rahul Shahi, and Mani Shankar. "Human computer interaction." 2010 3rd International Conference on Emerging Trends in Engineering and Technology. IEEE, 2010.

[65] Mousavi, Seyed Muhammad Hossein, et al. "Emotion Recognition in Adaptive Virtual Reality Settings: Challenges and Opportunities." WAMWB@ MobileHCI (2023): 1-20.

[66] Mason, Ian, Sebastian Starke, and Taku Komura. "Real-time style modelling of human locomotion via feature-wise transformations and local motion phases." Proceedings of the ACM on Computer Graphics and Interactive Techniques 5.1 (2022): 1-18.

[67] Song, Yang, et al. "Human motion analysis and measurement techniques: current application and developing trend." Analecta Technica Szegedinensia 17.2 (2023): 48-58.

[68] Hastie, Trevor, et al. The elements of statistical learning: data mining, inference, and prediction. Vol. 2. New York: springer, 2009.

[69] Breiman, Leo. "Random forests." Machine learning 45 (2001): 5-32.